\documentclass[journal]{IEEEtran}

\usepackage{float}
\usepackage{graphicx}
\usepackage[utf8]{inputenc}
\usepackage{amsmath,amssymb,amsfonts}
\usepackage{subcaption}
\usepackage{textcomp}
\usepackage{algorithmic}
\usepackage{graphicx}
\usepackage{url}
\usepackage{caption,setspace}
\usepackage{xcolor}
\usepackage{float}
\usepackage{multirow}
\usepackage{csquotes}
\usepackage{url}
\usepackage{tabularx}
\usepackage{booktabs}
\newcolumntype{Y}{>{\raggedleft\let\newline\\\arraybackslash\hspace{0pt}}X}
\newcolumntype{Z}{>{\centering\let\newline\\\arraybackslash\hspace{0pt}}X}

\usepackage{enumitem}
\setlist[itemize]{noitemsep,topsep=0pt}

\ifCLASSINFOpdf

\else

\fi

\begin{document}

\title{Single Morphing Attack Detection using Siamese Network and Few-shot learning}

\author{Juan Tapia,~\IEEEmembership{Member,~IEEE,}
        Daniel Schulz,
        and~Christoph Busch,~\IEEEmembership{Senior Member,~IEEE,}\\ 
   **The following paper is a pre-print. The publication is currently under review for IEEE.**     
\thanks{Juan Tapia and Christoph Busch, da/sec-Biometrics and Internet Security Research Group, Hochschule Darmstadt, Germany, e-mail: (juan.tapia-farias@h-da.de, christoph.busch@h-da.de).}
\thanks{Daniel Schulz, R+D Center, TOC Biometrics Company, Santiago, Chile, email: daniel.schulz@tocbiometrics.com}
\thanks{Manuscript received xxx; revised xx.}}

\markboth{Journal of \LaTeX\ Class Files,~Vol.~14, No.~8, August~2015}%
{Shell \MakeLowercase{\textit{et al.}}: Bare Demo of IEEEtran.cls for IEEE Journals}

\maketitle


\begin{abstract}
Face morphing attack detection is challenging and presents a concrete and severe threat for face verification systems. Reliable detection mechanism for such attacks, which have been tested with a robust cross-database protocol and unknown morphing tools still are a research challenge. This paper proposes a framework following the Few-Shot-Learning approach that shares image information based on the siamese network using triplet-semi-hard-loss to tackle the morphing attack detection and boost the clustering classification process. This network compares a bona fide or potentially morphed image with triplets of morphing and bona fide face images. Our results show that this new network cluster the data points, and assign them to classes in order to obtain a lower equal error rate in a cross-database scenario sharing only small image numbers from an unknown database. Few-shot learning helps to boost the learning process. Experimental results using a cross-datasets trained with FRGCv2 and tested with FERET and the AMSL open-access databases reduced the BPCER10 from 43\% to 4.91\% using ResNet50 and 5.50\% for MobileNetV2.
\end{abstract}


\maketitle

\section{Introduction}
Few-Shot Learning (FSL) aims for machine learning models to predict the correct class of samples when few examples are available in the training dataset~\cite{survey_FSL}. Using multiple domains allows the model to discover stable patterns across source domains, which generalise better to unseen domains.
This FSL approach can be broadly applied to the Morphing Attack Detection problem, when few examples are used to train a robust morphing attack detection system. This problem is even more relevant in a cross-dataset evaluation. This means to train with set $A$, validate in $A_i$ and test in set $B$, where $A$ and $B$ do not belongs to the same database, and $B$ is an unknown database. 

Morphing attack detection is a new topic aiming to detect unauthorised individuals who want to gain access to a "valid" identity in other countries. Morphing can be understood as a technique to combine two o more look-alike facial images from one subject and an accomplice, who could apply for a valid passport exploiting the accomplice's identity. Morphing takes place in the enrolment process stage. The threat of morphing attacks is known for border crossing or identification control scenarios. It can be broadly divided into two types: (1) Single Image Morphing Attack Detection (S-MAD) techniques (a.k.a as No-Reference MAD) and Differential Morphing Attack Detection (D-MAD) methods. The S-MAD scenario is more challenging as the decision needs to be taken on a single image without a trusted image available for the same subject~\cite{Venkatesh161}. 
S-MAD can be implemented different approaches: Textures, Shape, Quality, Hybrid features, Residual noise and Deep learning.

Siamese networks are a deep learning-based approach used in single and differential morphing attack detection. The siamese network takes bona fide and morphs pairs as input and yields a confidence score on the likelihood of using contrastive loss based on the euclidean distance of the pairs being from the same subject. Siamese networks are ideal for this purpose as they are primarily employed in tasks that require finding similarities between two inputs.

Most of the time, traditional approaches fail with new examples from an unknown database because the contrastive loss assigns the new samples based only on euclidean distance. A new morph image will look like a bona fide image. Then, it will be assigned erroneously to bona fide class even when it is a morphed image. We explored a triplet hard-loss and semi-hard loss to improve this limitation. 

This work proposes to use a triplet-loss function to estimate the difficulties of each morphing tool and examine different pre-trained networks as a general framework solution. The experiments are conducted with three databases: FERET, FRGCv2 and the open-source AMSL Morph. Each one of them presents subsets of different morphing tools and conditions such as resize, print and scan.
As an essential contribution, a FSL is used to include only an small amount of examples from a new unknown dataset to guide the training process and increase the performance of the method. These few examples allow us to cluster the different morphing tools and the attacks.

The summary contributions of this work are described as follows: 
\begin{itemize}
    \item A modified siamese network is proposed based on triplet semi-hard loss.
    \item A FSL approach is proposed to guide and boost the learning process to improve the results.
    \item A cross-database evaluation is performed, observing a reduction of the Equal Error Rate.
    \item A visualisation study based on t-Distributed Stochastic Neighbour Embedding (t-SNE)~\cite{tsne} is presented in order to understand how the morphed and bona fide images are clustered after few-shot-learning was applied.
    \item This approach is a general-purpose framework and can be extended to other domains.
\end{itemize}

This paper is organised as follows: The related work is described in Section~\ref{sec:related}. The method is outlined in Section~\ref{method}. The database are introduced in section~\ref{database} and the experiments and results are presented in section \ref{exp}. Finally, conclusion are presented in section~\ref{conclusion}.

\section{Related work}\label{sec:related}

Face morphing attack has captured the interest of the research community and government agencies in globally but more specifically in Europe. For instance the European Union (EU) decided to fund the image Manipulation Attack Resolving Solutions (iMARS) project \footnote{\url{https://cordis.europa.eu/project/id/883356}}, developing new techniques for detection of morphed images.

Ferrera et al.~\cite{Ferrara} were the first to investigate the face recognition system's vulnerability with regards to morphing attacks. They has evaluated the feasibility of creating deceiving morphed face images and analysed the robustness of commercial face recognition systems in the presence of morphing.

Scherlag et al.~\cite{Scherhag_survey} provided and developed a survey about the impact of morphing images on face recognition systems. The same authors~\cite{ScherhagDeep} proposed a face representation from embedding vectors for differential morphing attack detection, creating a more realistic database, different scenarios, and constraints with four automatic morphed tools. They also reported detection performances for several texture descriptors in conjunction with machine learning techniques.

Damer et al.~\cite{pixelwise} proposed a pixel-wise supervision approach where they train a network to classify each pixel of the image into an attack or not, rather than only having one label for the whole image. Their Pixel-Wise Morphing Attack Detection (PW-MAD) solution proved to perform more accurately than a set of established baselines.

Tapia et al.~\cite{TapiaSMAD} proposed to add an extra stage of feature selection after feature extraction based on Mutual Information to estimate and keep the most relevant and remove the most redundant features from the face images to separate bona fide and morphed images on a S-MAD scenario. The high redundancy between features confuses the classifier. This approach visualised the most relevant features and identified which areas are the most sensible to detect morphed images. The authors conclude that the eyes and nose are the most relevant areas.

Most of the state-of-the-art approaches use machine learning and deep learning to detect and classify morph images. Also, they are utilising embedding vectors from deep learning approaches to detect and classify the images. Our paper focused on the Deep learning method. 

In the machine learning scenario, popular texture-based methods include local binary patterns (LBPs)~\cite{Raghavendra,ScherhaglBP}, Local Phase Quantisation~(LPQ) features~\cite{Ojansivu} and Binarised Statistical Image Features~(BSIF)~\cite{bsif-kanala, RaghavendraMAFI, Raghavendra81} has been used. Furthermore, these texture features have also been extracted for different colour channels to obtain a robust detection performance and fusion~\cite{Siri}. Several variants of LBPs and a histogram of oriented gradients (HOG) have also been explored in the literature~\cite{SchergagHOG}. 

Regarding to deep learning, some work has been previously proposed based on siamese network.

Borgui et al.~\cite{siamese-guido} proposed a differential morph attack detection based on a double Siamese architecture. The pre-processing of input images, and the adopted training. The proposed framework consisted of two different modules, referred to as “Identity” and “Artifact” blocks, respectively, and each block was based on a siamese network followed by a Multi-Layer Perceptron (MLP) that act as fusion layers. Finally, a Fully Connected layer (FC) merges the features originated from the two modules and outputs the final score. Experimental results were obtained in three datasets, namely PMDB, MorphDB, and AMSL. This approach used a Contrastive loss.

Soleymani et al.~\cite{Soleymani} propose to develop a novel differential morphing attack detection algorithm using a deep siamese network. The siamese network takes image pairs as inputs and yields a confidence score on the likelihood that the face images are
from the same subject. They employ a pre-trained Inception ResNetv1 as the base network initialised with weights pre-trained on VGGFace2. The experiments are conducted on two separate morphed image datasets: VISAPP17 and MorGAN. This approach used a contrastive loss.

Chaudhary et al.~\cite{Chaudhary_2021_CVPR} proposed a differential morph attack detection algorithm using an undecimated 2D Discrete Wavelet Transform (DWT). By decomposing an image to wavelet sub-bands, we can identify the morph artefacts that are hidden in the image domain more clearly in the spatial frequency domain. 

It is essential to highlight, that the NIST Face Recognition Vendor Test (FRVT) with the specialisation on MAD~\cite{Ngan2020FaceRV} has been used to evaluate and report the performances of different morph detection algorithms~\cite{Soler, Siri}. This task is recommended but not a mandatory requirement.
The test is organised in three tiers according to the morph images quality. Tier 1 evaluates low-quality morph images; Tier 2 considers automatic morph images; and Tier 3 for high-quality images. Further, the NIST report is organised w.r.t local (crop faces) and global (passport-photos) morphing algorithms. This test show the relevance of the the morphing attack detection research.

\section{Method}
\label{method}

In a classical classification problem, features are learned to maximise inter-class distances between classes and minimise intra-class variances within categories. However, such a discriminative approach is challenging  without multiple samples per class. 
In order to improve this limitation, a few-shot-learning approach is proposed based on a siamese networks and domain generalisation~\cite{DomainGI}, that takes into account that we have only a few examples ($N$) of the new classes (morphing images generated with different tools). The optimal number of examples $N$ must be determined. Different loss-functions are also explored to determine the best way to assign the class to examples belonging to each new morphing tool. 

\begin{figure*}[]
\centering
	\includegraphics[scale=0.40]{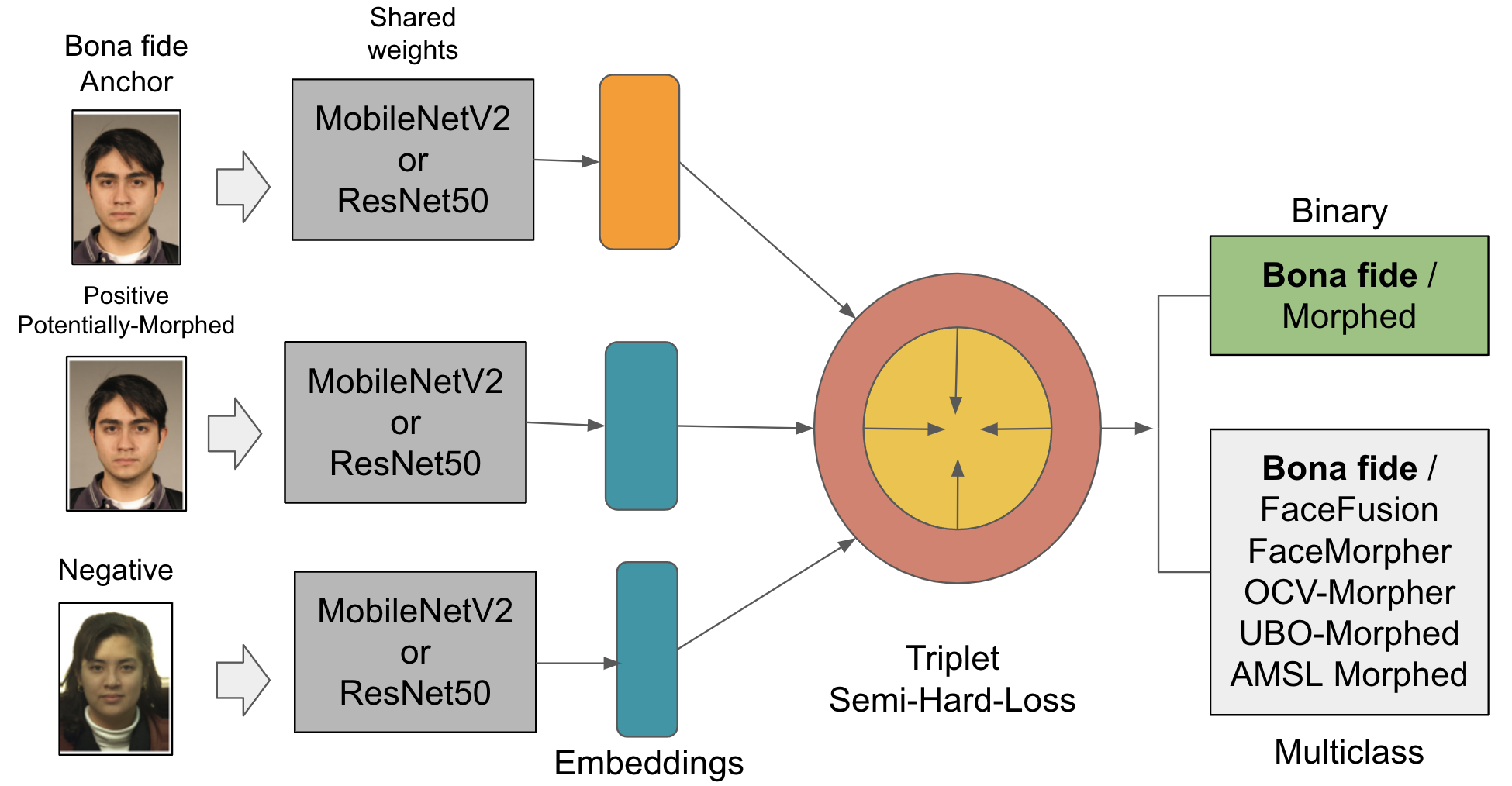}
\caption{General framework representation, proposed with a modified siamese network for a Morphing attack detection.}
\label{fig:framwork}
\end{figure*}

Figure~\ref{fig:framwork} shows the proposed framework used in this paper, where an anchor, a positive and a negative images are used to evaluate with a semi-hard-loss function a siamese network.

\subsection{Siamese Network}

The siamese network was first proposed by Bromley et al.~\cite{bromley1993signature}, aimed to the signature verification problem. It consists of two identical networks, which can use different inputs, joined at their outputs by an energy function, that calculates a metric between the features estimated by each network. This kind of network is ideal for morphing attack detection, as they are primarily designed to find similarities between two inputs. 

Tipically, contrastive loss is used as the loss function for training a siamese network, to distinguish between genuine (non-morphed) and impostor (morphed) pairs. Traditionally, the contrastive loss function optimises two identical DNNs outputs, each operating on a different input image and using a euclidean distance measure or an SVM classifier to make the final decision. While contrastive representations have achieved state-of-the-art performance on visual recognition tasks and have been theoretically proven to be effective for binary classification, we argue that it could be problematic for multi-class classification. 

In order to extract features for the bona fide and morphed images, ImageNet pre-trained general-purpose backbones such as ResNet50~\cite{he2016deep} and MobileNetV2~\cite{sandler2018mobilenetv2} were used. Then, the network is fine-tuned with a morphing database. The model is optimised by enforcing the loss function to separate both classes in a binary problem or $N$ classes in a multi-class problem. 

Regarding the analysis of this problem in order to improve the results, we can explore a binary approach, which means separating into two classes bona fide and morphed by distances with a low number of new samples. On the other hand, we can use a few-shot learning approach to assign a label to each sample. Then, this is a multiclass problem in our case bona fide, versus different kinds of morphing tools such as FaceFusion, Face-Morpher, Opencv-Morpher, UBO-Morpher and others.

Our classifier is evaluated using metrics according to ISO-30107-3 \footnote{\url{https://www.iso.org/standard/67381.html}}. The details are explained in section~\ref{sec:metric}.

\subsection{Loss function}

As we mentioned before, traditionally convolutional neural networks used a soft-max function with a cross-entropy loss because we have a fixed number of classes. However, in some cases, we need to have a variable number of classes. In face recognition, for instance, we need to be able to compare two unknown faces images and determine whether they are from the same subject or not \cite{SchroffKP15}. This approach seems to be useful for the Morphing Attack Detection task, so we evaluated losses used in face recognition with siamese networks.

\subsubsection{Contrastive Loss}
\label{CL}

It is a distance-based contrastive loss, as opposed to more conventional error-prediction losses. This loss is used to learn embeddings in which two images belonging to the same class have a low euclidean distance, and two images belonging to different classes have a large euclidean distance. The contrastive loss $L_c$ is defined as follows:

\begin{equation}
    L_c= (1-y_g)d(I_1, I_2)^2 + y_g (\max(0, m - d(I_1,I_2))^2
\end{equation}
where $d(I_1, I_2)$ is the distance in the embedding space between two input images, $m$ is the margin or distance threshold to control the separation, and $y_g$ is 0 for image pairs with the same label, and $y_g$ is 1 for pairs with different labels.




\subsubsection{Triplet Loss}
\label{TL}

It is a loss function that trains a neural network to closely embedded features of the same class while maximising the distance between embeddings of different classes. An anchor is chosen along with one negative and one positive sample to do this~\cite{SchroffKP15}.

To formalise this requirement, a loss function is defined over triplets of embeddings:

\begin{itemize}
    \item An anchor image (a) - bona fide.
    \item A positive image of (p) the same class as the anchor - bona fide.
    \item A negative image (n) of a different class - morphed.
\end{itemize}

For some distance $(d)$ on the embedding space  the loss of a triplet $(a,p,n)$ is defined as:
\begin{equation}
    L_t = \max(d(a,p)-d(a,n)+margin,0)
\end{equation}

We minimise this loss, which pushes $d(a,p)$ to $0$ and $d(a,n)$ to be greater than $d(a,p)+margin$. As soon as $n$  becomes an “easy negative”, the loss becomes zero.

Based on the definition of the loss, there are three categories of triplets:
\begin{itemize}
    \item Easy triplets: triplets which have a loss of 0, because $d(a,p)+margin<d(a,n)$
    \item Hard triplets: triplets where the negative is closer to the anchor than the positive, i.e. $d(a,n)<d(a,p)$
     \item Semi-hard triplets: triplets where the negative is not closer to the anchor than the positive, but which still have positive loss: $d(a,p)<d(a,n)<d(a,p)+margin$.
\end{itemize}

Each of these definitions depend on "where" the negative (morph image) is localised, relatively to the anchor and positive (bonafide). We can therefore extend these three categories to the negatives: hard negatives, semi-hard negatives or easy negatives. 






\section{Databases}
\label{database}

The FERET and FRGCv2 databases were used to create the morph images based on the protocol described in~\cite{ScherhagDeep}. The AMSL Face Morph open-access image dataset was also used to evaluate the best algorithm. A summary of the databases is presented in Table~\ref{tab:database}. All the images were captured in a controlled scenario and include variations in pose and illumination.
FRGCv2 presents images more compliant to the passport portrait photo requirements. The images contain illumination variation, different sharpness and changes in the background. 
The original images have a size of $720\times960$ pixels. For this paper, different scenarios of post processing were used such as No Post Processing (NPP), Print and Scan with 300 dpi (PS300) and Print and Scan with 600 dpi (PS600). The faces were detected, and images were resized and reduced to $360 \times 480$ pixels. These images still fulfill the resolution requirement of the intra-eye distance of 90 pixels defined by ICAO-9303-p9-2015.

The AMSL Face Morph image open-access dataset was created based on images from Face Research Lab London (London DB) set\footnote{\url{https://figshare.com/articles/dataset/Face_Research_Lab_London_Set/5047666}}. This dataset includes genuine neutral and smiling faces as well as morphed face images. Also, all the images are ICAO compliant. The images were downscaled to $531 \times 413$ pixels. These images present high-quality face morphed images \cite{neubert58}.

The $\alpha$ value to define the contribution of each subject to morph image results was 0.5 for all the morph images. 
The morphing tools used are described in Table~\ref{tab:database_morph} per scenario.

\begin{table}[H]
\scriptsize
\centering
\caption{Number (Nº) of images used for FERET, FRGCv2 Database and Face Research London dataset. Column 1, show the name of the dataset.}
\begin{tabular}{lllll}
\hline
\textbf{Database} & \textbf{Nº Subjects} & \textbf{Bona fide} & \textbf{Morphs} & \\ \hline
FRGCv2            & 533              & 984               & 964             &            \\ \hline
FERET             & 529              & 529               & 529             &             \\ \hline
Face Res. London DB & 102            & 102              & 2,175         &            \\ \hline
\end{tabular}
\label{tab:database}
\end{table}

\begin{table}[H]
\scriptsize
\centering
\caption{Morphing tool softwares and number of images created by each method. The number of images is per dataset (FRGCv2/FERET).}
\begin{tabular}{lllll}
\hline
\textbf{Database} & \textbf{Nº Subjects} & \textbf{Bona fide} & \textbf{Morphs} &  \\ \hline
FaceFusion        & 533/529              & 984/529               & 964/529             &             \\ \hline
FaceMorpher       & 533/529              & 984/529               & 964/529             &            \\ \hline
OpenCV-Morph        & 533/529              & 984/529               & 964/529          &             \\ \hline
UBO-Morpher       & 533/529              & 984 /529              & 964/529             &            \\ \hline
AMSL-FaceMorph   & 102                   & 204                   & 2,175             &             \\ \hline
\end{tabular}
\label{tab:database_morph}
\end{table}

The following algorithms were used to create morph images:
\begin{itemize}
    \item FaceFusion is a proprietary morphing algorithm, developed as an iOS app \footnote{\url{www.wearemoment.com/FaceFusion/}}. This algorithm to create high-quality morph images without visible artefacts.
    \item FaceMorpher is an open-source algorithm to create morph images \footnote{\url{https://github.com/alyssaq/face_morpher}}. This algorithm introduce some artefacts in the background.
    \item OpenCV-Morph, this algorithm is based on the OpenCV implementation \footnote{\url{www.learnopencv.com/face-morph-using-opencv-cpp-python}}. The images contain visible artefacts in the background and some areas of the face.
    \item Face UBO-Morpher\cite{UBO}. The University of Bologna developed this algorithm. The resulting images are of high quality without artefacts in the background. 
    \item AMSL-FaceMorph\cite{neubert58}. The images were created on Face Research Lab London Set with an in-house method. The resulting images are of high quality without artefacts in the background. 
\end{itemize}

\begin{figure*}[!h]
\includegraphics[width=\linewidth]{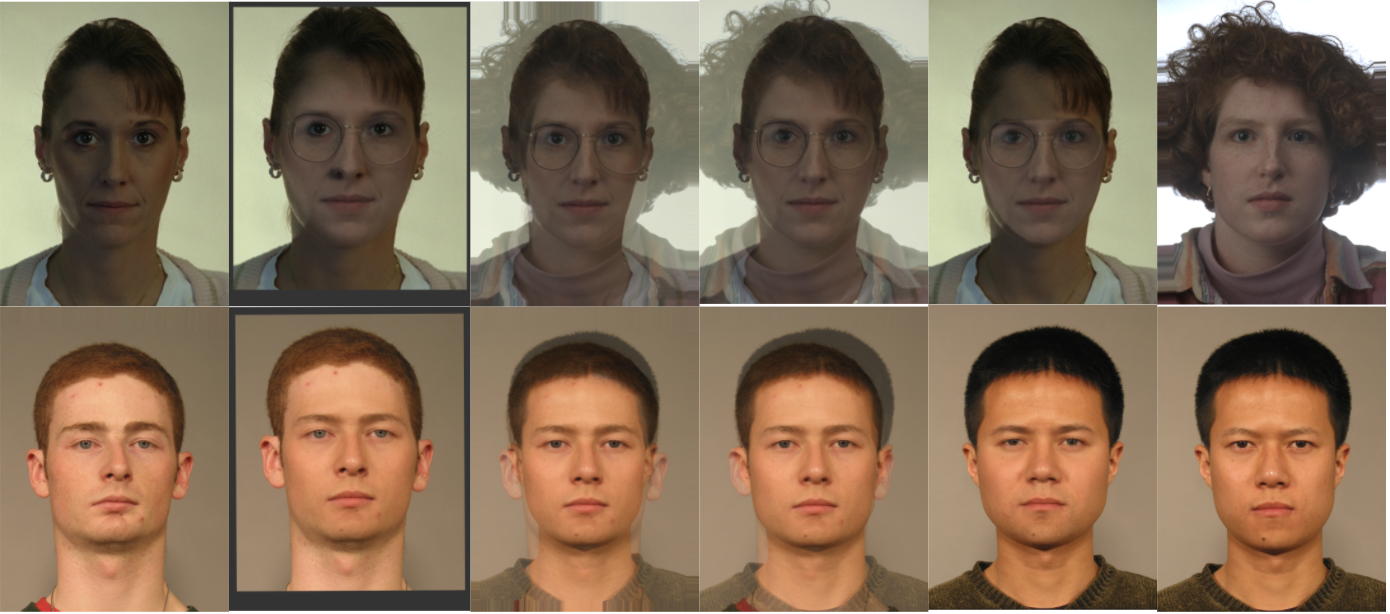}
\caption*{
Subject 1 \ \ \ \ \ \ \ \ \ \ \ \ \ \ \ \ \ \ \ \ \ \ \,  
FaceFusion \ \ \ \ \ \ \ \ \ \ \ \ \ \ \ \ \ \ \ \ \,
FaceMorpher \ \ \ \ \ \ \ \ \ \ \ \ \ \ \ \ \ \ \,  
Opencv-Morpher \ \ \ \ \ \ \ \ \ \ \ \ \ \ \ \ \ \,
UBO-Morpher \ \ \ \ \ \ \ \ \ \ \ \ \ \ \ \ \ \ \,  
Subject2}

\caption{Examples of different morphing algorithms for two subjects in the FERET and FRGCv2 databases.}
\label{fig:datasetex}
\end{figure*}

Figure \ref{fig:datasetex} show examples of the morphing portrait images and the different output qualities with artefacts in their background. For instance, Opencv-Morpher show a lot of noise in a portrait implementation. 

\begin{figure}[H]
\centering
	\includegraphics[scale=0.30]{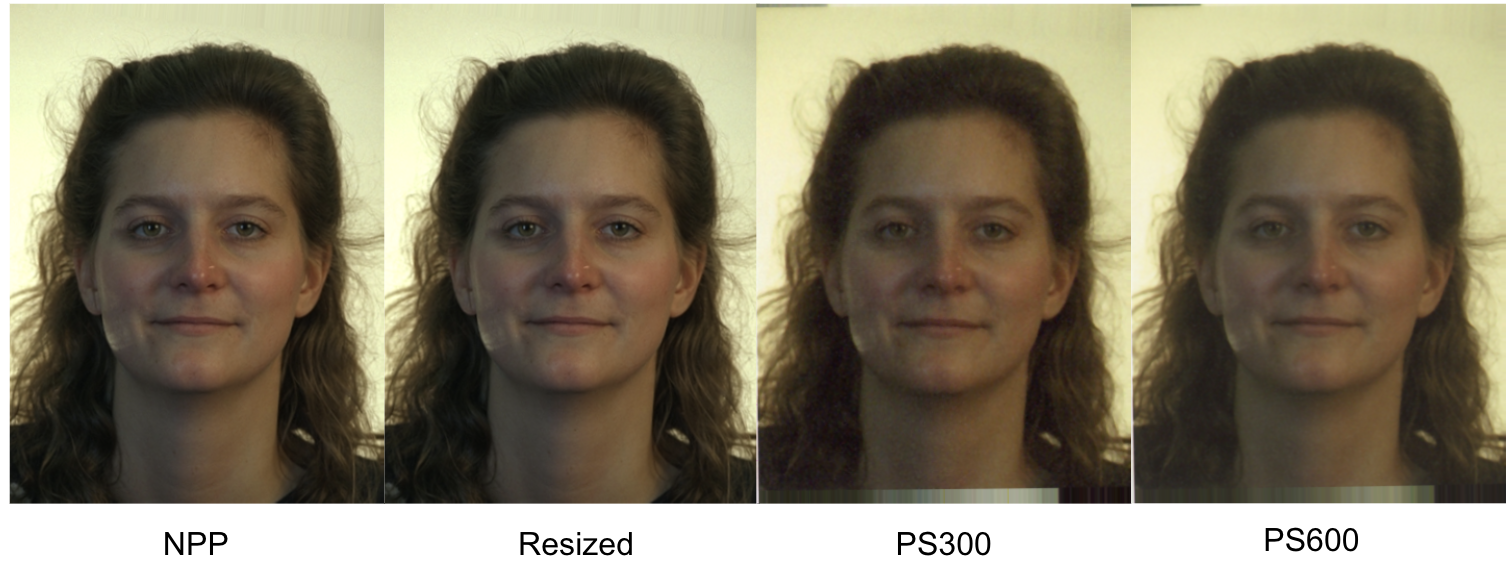}
\caption{Examples of bona fide FERET images from the four different scenarios.}
\label{fig:quality}
\end{figure}

Figure \ref{fig:quality} show examples of different scenarios, No Post Processing (NPP), Resized (NPP reduced by half size), Print and Scan with 300 dpi (PS300), and Print and Scan with 600 dpi (PS600). 

As we mentioned before, after creation of the morphed images, all the faces were cropped using a modified dlib face detector implementation \footnote{https://www.pyimagesearch.com/2018/09/24/opencv-face-recognition/}. 

\section{Metrics}
\label{sec:metric}

The ISO/IEC 30107-3 standard\footnote{\url{https://www.iso.org/standard/67381.html}} presents methodologies for the evaluation of the performance of PAD algorithms for biometric systems. The APCER metric measures the proportion of attack presentations---for each different PAI---incorrectly classified as bona fide (genuine) presentations. This metric is calculated for each PAI, where ultimately the worst-case scenario is considered. Equation~\ref{eq:apcer} details how to compute the APCER metric, in which the value of $N_{PAIS}$ corresponds to the number of attack presentation images, where $RES_{i}$ for the $i$th image is $1$ if the algorithm classifies it as an attack presentation (morphed image), or $0$ if it is classified as a bona fide presentation (real image).

\begin{equation}\label{eq:apcer}
    {APCER_{PAIS}}=1 - (\frac{1}{N_{PAIS}})\sum_{i=1}^{N_{PAIS}}RES_{i}
\end{equation}

Additionally, the BPCER metric measures the proportion of bona fide presentations mistakenly classified as attacks presentations to the biometric capture device, or the ratio between false rejection to total genuine attempts. The BPCER metric is formulated according to equation~\ref{eq:bpcer}, where $N_{BF}$ corresponds to the number of bona fide presentation images, and $RES_{i}$ takes identical values of those of the APCER metric.

\begin{equation}\label{eq:bpcer}
    BPCER=\frac{\sum_{i=1}^{N_{BF}}RES_{i}}{N_{BF}}
\end{equation}

These metrics effectively measure to what degree the algorithm confuses presentations of morphed images with bona fide images, and vice versa. The APCER, BPCER are dependent on a decision threshold.

A Detection Error Trade-off (DET) curve is also reported for all the experiments. In the DET curve, the Equal Error Rate (EER) value represents the trade-off when the APCER is equal to the BPCER. Values in this curve are presented as percentages. Additionally, two different operational points are reported, according to ISO/IEC 30107-3. BPCER\textsubscript{10} which corresponds to the BPCER when the APCER is fixed at 10\%, and BPCER\textsubscript{20} which is the BPCER when the APCER is fixed at 5\%. BPCER\textsubscript{10} and BPCER\textsubscript{20} are independent of decision thresholds.

\section{Experiments and Results} 
\label{exp}

This section explores the quantitative results of the proposed scheme based on siamese network and three loss functions: contrastive loss, semi-hard triplet loss and hard triplet loss. 

All the experiments were performed using a cross-database protocol, with FRGCV2 for training and FERET for test, and vice versa.

The FERET and FRGCv2 databases were partitioned to have 60\% training and 40\% testing data. The London DB was used only for test as an unknown cross-dataset scenario.

In total 12,000 images were used for training and 2,500 for validation from the FRGCv2 database. 
For testing, 2,000 FERET images were used, distributed in NPP (500 images), Resize (500 images), PS300 (500 images), PS600 (500 images). All the subsets contains: bona fide (100 images), FaceFusion (100 images), FaceMorpher (100 images), Opencv-Morpher (100 images) and UBO-Morpher (100 images). AMSL test set contains 2,175 morphed images and 204 bona fide images.

For inference, let $f(x, w)$ be the embedding function, parameterised by a neural network and $\varphi(x_1, x_2, w)$ the
euclidean distance between points $x_1$ and $x_2$ in the embedding space given by $w$:
\begin{equation}\label{eq:dist}
    \varphi(x_1, x_2, w) = \parallel f(x_1, w)-f(x_2, w) \parallel ^{2}
\end{equation}
In our case, we calculate the euclidean distance between an input image $x$ and each bona fide template $t_i \in T = \{t_1, ..., t_N\}$, then, we average all the distances:
\begin{equation}\label{eq:avg}
    \varphi_{avg} = \frac{1}{N}\sum_{i=1}^{N}{\varphi(x, t_i, w)}
\end{equation}
Finally, we compare the average distance $\varphi_{avg}$ with a threshold $th$, when a lower distance indicates a greater likelihood of the bona fide class. The threshold must be chosen according to the desired operational point.
\begin{equation}\label{eq:thresh}
    y = \begin{cases}
            1 & \text{if } \varphi_{avg} < th\\
            0 & \text{if } \varphi_{avg} \geqslant th\\
         \end{cases}
\end{equation}

\begin{figure*}[]
\centering
	\includegraphics[scale=0.45]{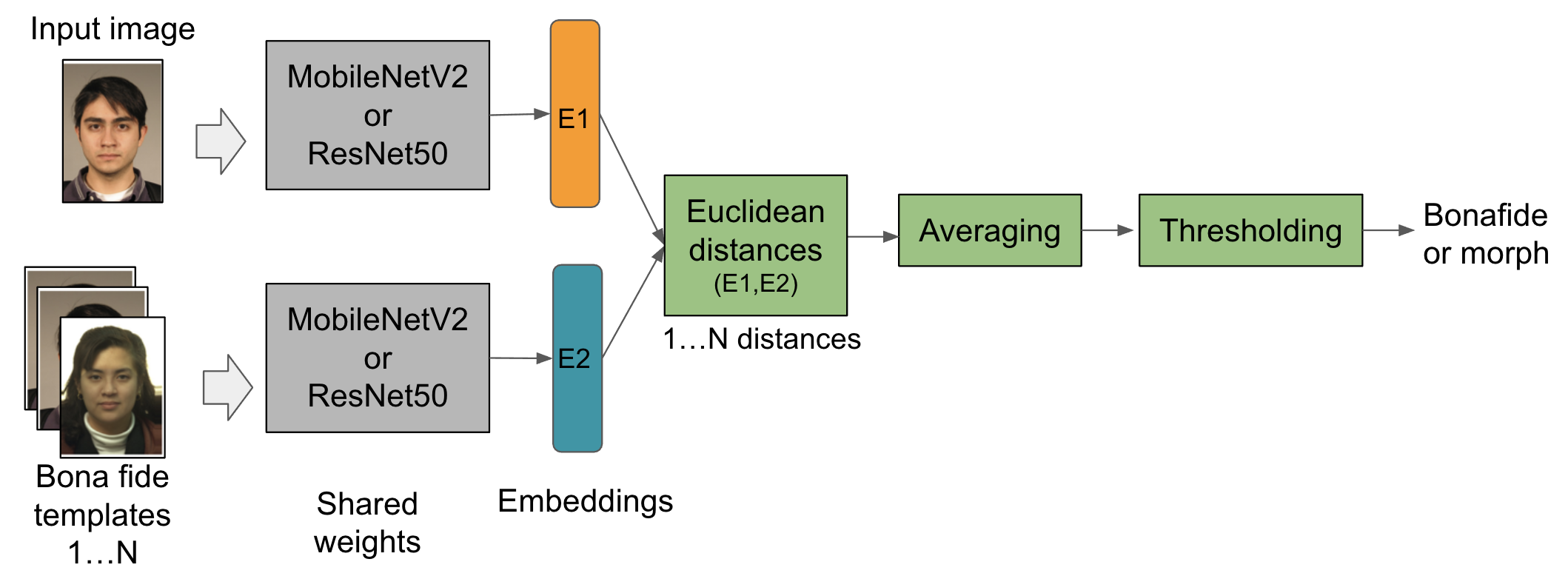}
\caption{Representation of inference time workflow for siamese network.}
\label{fig:inference}
\end{figure*}

For the template images, currently are sampled in a random fashion, but it must be calculated the optimal number of samples. For this purpose, a grid search using 1, 2, 4, 8 and 12 samples was performed. Table \ref{tab:templates} show the grid search results, where best result was reached using four random templates, showing an appropriate speed vs. performance trade-off. All the inference process is depicted in Figure \ref{fig:inference}.

\begin{table}[H]
\centering
\scriptsize
\caption{Selection of the best number of templates for siamese network.} 

\label{tab:templates}
\begin{tabular}{|c|c|c|c|}
\hline
Model                     & \begin{tabular}[c]{@{}c@{}}N\\ Templates\end{tabular} & \begin{tabular}[c]{@{}c@{}}D-EER\\ (\%)\end{tabular} & \begin{tabular}[c]{@{}c@{}}BPCER10\\ (\%)\end{tabular} \\ \hline
\multirow{5}{*}{ResNet50} & 1                                                     & 18.23                                                & 21.03                                                  \\ \cline{2-4} 
                          & 2                                                     & 15.33                                                & 19.23                                                  \\ \cline{2-4} 
                          & \textbf{4}                                                     & \textbf{12.10}                                                & \textbf{13.01}                                                  \\ \cline{2-4} 
                          & 8                                                     & 13.25                                                & 15.23                                                  \\ \cline{2-4} 
                          & 12                                                    & 15.13                                                & 17.23                                                  \\ \hline
\end{tabular}
\end{table}


\subsection{Data-Augmentation (DA)}
In general, solving an FSL problem by augmenting the data is straightforward to understand. The data is augmented by taking advantage of the preliminary information for the target task. However, the weakness of solving the FSL problem by data augmentation is that the augmentation policy is often tailor-made for each data set ad-hoc and cannot be used easily on other data sets created for a different morphing tool (especially for data sets from different domains). 
Then, three different DA levels were explored. Light, Medium, Heavy. The Heavy augmentation reached the best results with the followings operation based on imgaug library~\cite{imgaug}. Then, the results only with aggressive DA are reported:
\begin{itemize}
    \item Affine\_chance=0.5
    \item Crop\_chance=0-0.5
    \item Brightness\_chance=0.5
    \item Temperature\_chance=0.35
    \item Blur\_chance=0.5,
    \item Noise\_chance=0.35, 
    \item Dropout\_chance=0-0.35,
    \item Random\_order=False.
    
\end{itemize}

\subsection{Experiment 1}
This experiment was developed as a baseline. First, FRGCv2 are used to train and test a siamese network. Afterwards, the FRGCv2 was used to train and the test was performed in FERET. Both datasets are subject-disjoint. All models were trained using a ResNet50 backbone, pre-trained with ImageNet ~\cite{russakovsky2015imagenet}, with all the layer being trainable.

As expected, in Table~\ref{tab:exp1}~(columns 3 and 4) the D-EER and BPCER10 are presented, obtaining very low values, as stated in the literature.  This confirm the fact, that methods evaluated on images belonging to the same dataset, even being person-disjoint, are not real challenging. 
Conversely, when a cross-database protocol was used, the results obtained in D-EER and BPCER10 are worse for a large margin, as shown in Table~\ref{tab:exp1}~(columns 5 and 6).

This experiment was also used to evaluate how the three proposed loss functions performed, with Triplet loss Semi-Hard~(SHL) obtaining the best results. Then, Our goal is to reduce BPCER10 of 43.00\% to a value lower than 10\% as reached with system trained and tested with the same database. The results are shown in Table~\ref{tab:exp1}. 

\begin{table}[H]
\scriptsize
\centering
\caption{Results experiment 1. CL: represents Constrastive Loss. SHL, Triplet Loss Semi-Hard, HL, Triplet Loss Hard. }
\label{tab:exp1}
\begin{tabular}{|c|ccc|cc|}
\hline
\multirow{2}{*}{Model} & \multicolumn{3}{c|}{\begin{tabular}[c]{@{}c@{}}Train:FRGCv2 / \\ Test: FRGCv2\end{tabular}}                                                                    & \multicolumn{2}{c|}{\begin{tabular}[c]{@{}c@{}}Train: FRGCv2 /\\ Test: FERET\end{tabular}}                                         \\ \cline{2-6} 
                       & \multicolumn{1}{c|}{Loss} & \multicolumn{1}{c|}{\begin{tabular}[c]{@{}c@{}}D-EER\\ (\%)\end{tabular}} & \begin{tabular}[c]{@{}c@{}}BPCER10\\ (\%)\end{tabular} & \multicolumn{1}{c|}{\begin{tabular}[c]{@{}c@{}}D-EER\\ (\%)\end{tabular}} & \begin{tabular}[c]{@{}c@{}}BPCER10\\ (\%)\end{tabular} \\ \hline
ResNet50               & \multicolumn{1}{c|}{CL}   & \multicolumn{1}{c|}{12.40}                                                & 16.60                                                  & \multicolumn{1}{c|}{31.90}                                                & 62.10                                                  \\ \hline
\textbf{ResNet50}               & \multicolumn{1}{c|}{\textbf{SHL}}  & \multicolumn{1}{c|}{\textbf{12.10}}                                                & \textbf{13.01}                                                  & \multicolumn{1}{c|}{\textbf{21.10}}                                                & \textbf{43.00}                                                  \\ \hline
ResNet50               & \multicolumn{1}{c|}{HL}   & \multicolumn{1}{c|}{38.30}                                                & 13.10                                                  & \multicolumn{1}{c|}{43.10}                                                & 65.00                                                  \\ \hline
\end{tabular}
\end{table}

\subsection{Experiment 2}
This experiment was developed to explore how the triplet loss semi-hard function, selected in the previous section, performs fine-tuning ImageNet pre-trained backbones. As a main difference compared to experiment 1, only the cross-database protocol was implemented. FRGCv2 and FERET datasets were used, training on the first, testing on the second. Results are shown in Table~\ref{tab:exp2}. We also performed experiments freezing layers in the network. 

For MobileNetV2, we freeze all the layers up to the following: block\_1\_expand\_1 and block\_1\_expand\_2. 

For ResNet50, all the layers were freeze up to: conv2\_block1\_1\_conv and conv5\_block1\_1\_conv. The best results was obtained with ResNet50 and Conv2\_block1\_1 with a EER of 9.39\%.

\begin{table}[H]
\scriptsize
\centering
\caption{Results experiment 2.  Semi Hard Loss, Heavy data augmentation. LR: represents Learning Rate.}
\label{tab:exp2}
\begin{tabular}{|c|llccc|}
\hline
\multirow{2}{*}{Model} & \multicolumn{5}{c|}{\begin{tabular}[c]{@{}c@{}}Train: FRGCv2 /\\ Test: FERET\end{tabular}}                                                                                                                                                                                           \\ \cline{2-6} 
                       & \multicolumn{1}{c|}{Layers}           & \multicolumn{1}{l|}{LR}   & \multicolumn{1}{c|}{\begin{tabular}[c]{@{}c@{}}D-EER\\ (\%)\end{tabular}} & \multicolumn{1}{c|}{\begin{tabular}[c]{@{}c@{}}BPCER10\\ (\%)\end{tabular}} & \begin{tabular}[c]{@{}c@{}}BPCER20\\ (\%)\end{tabular} \\ \hline
MobileNetV2            & \multicolumn{1}{l|}{Blk1\_2\_expand}     & \multicolumn{1}{l|}{2e-5} & \multicolumn{1}{c|}{21.10}                                               & \multicolumn{1}{c|}{24.10}                                                  & 25.40                                                  \\ \hline
MobileNetV2            & \multicolumn{1}{l|}{Blk1\_1\_expand}    & \multicolumn{1}{l|}{2e-5} & \multicolumn{1}{c|}{11.10}                                                & \multicolumn{1}{c|}{11.50}                                                  & 20.50                                                  \\ \hline
MobileNetV2            & \multicolumn{1}{l|}{Blk1\_2\_expand}    & \multicolumn{1}{l|}{2e-5} & \multicolumn{1}{c|}{15.48}                                               & \multicolumn{1}{c|}{12.60}                                                  & 13.00                                                  \\ \hline
ResNet50               & \multicolumn{1}{l|}{Conv2\_block1\_1} & \multicolumn{1}{l|}{1e-4} & \multicolumn{1}{c|}{9.50}                                               & \multicolumn{1}{c|}{9.49}                                                 & 28.50                                                 \\ \hline
\textbf{ResNet50}               & \multicolumn{1}{l|}{\textbf{Conv2\_block1\_1}} & \multicolumn{1}{l|}{\textbf{1e-5}} & \multicolumn{1}{c|}{\textbf{9.39}}                                               & \multicolumn{1}{c|}{\textbf{9.49}}                                                 & \textbf{18.74}                                                 \\ \hline
ResNet50               & \multicolumn{1}{l|}{Conv5\_block1\_1} & \multicolumn{1}{l|}{1e-4} & \multicolumn{1}{c|}{14.50}                                               & \multicolumn{1}{c|}{22.25}                                                 & 41.50                                                 \\ \hline
ResNet50               & \multicolumn{1}{l|}{Conv5\_block1\_1} & \multicolumn{1}{l|}{1e-3} & \multicolumn{1}{c|}{13.24}                                               & \multicolumn{1}{c|}{16.50}                                                 & 29.50                                                 \\ \hline
\end{tabular}
\end{table}

\begin{figure*}[]
\centering
	\includegraphics[scale=0.22]{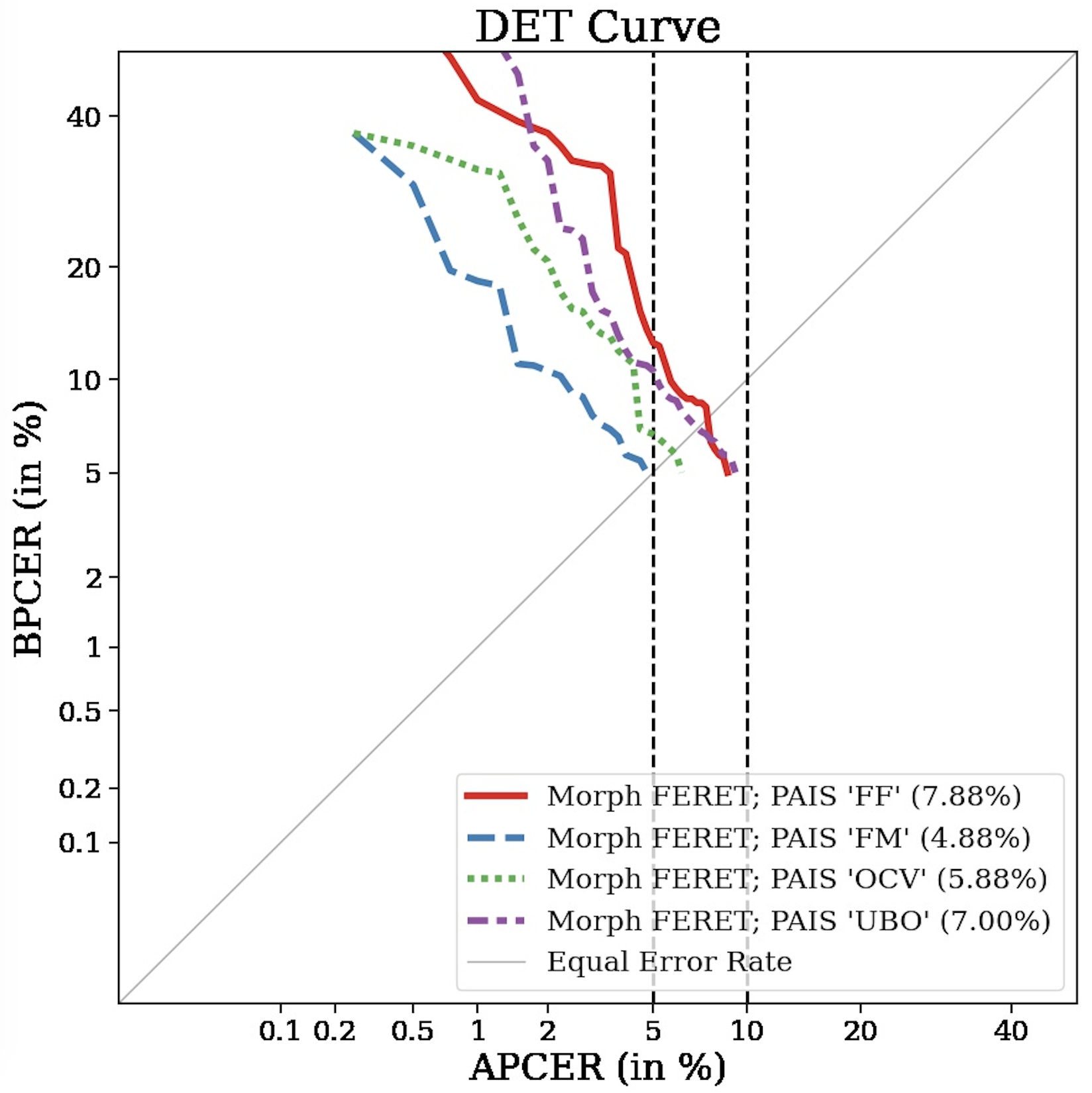}
	\includegraphics[scale=0.285]{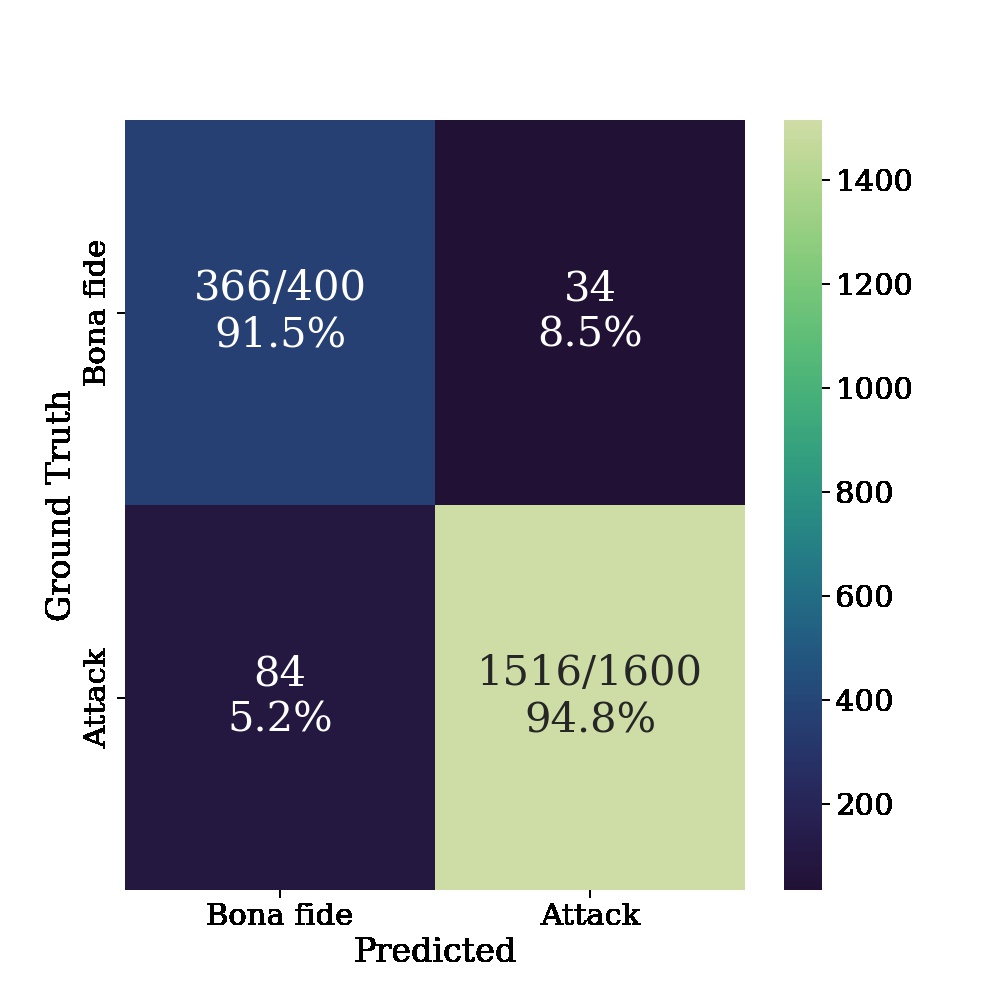}\\
	\includegraphics[scale=0.22]{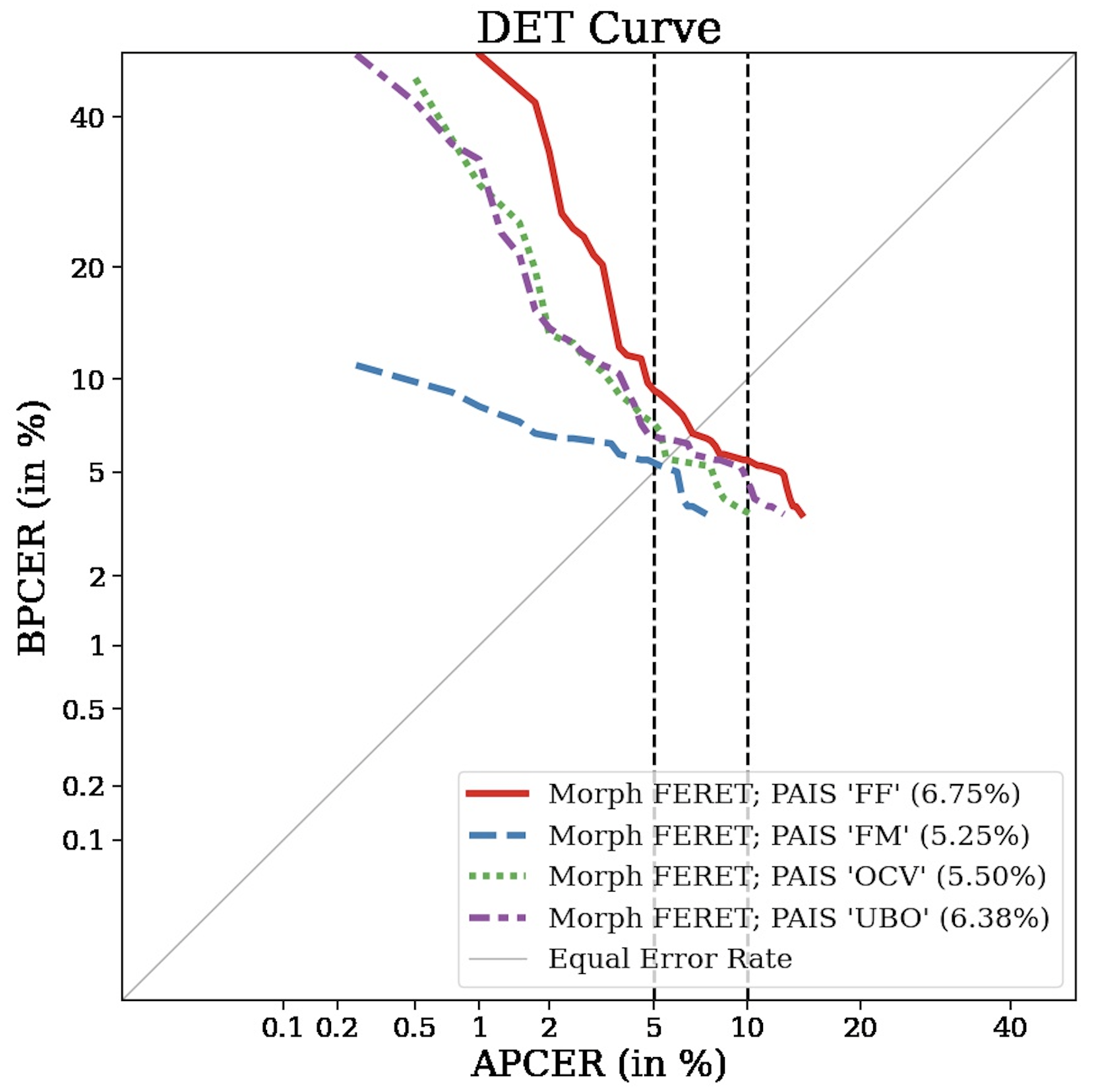}
	\includegraphics[scale=0.285]{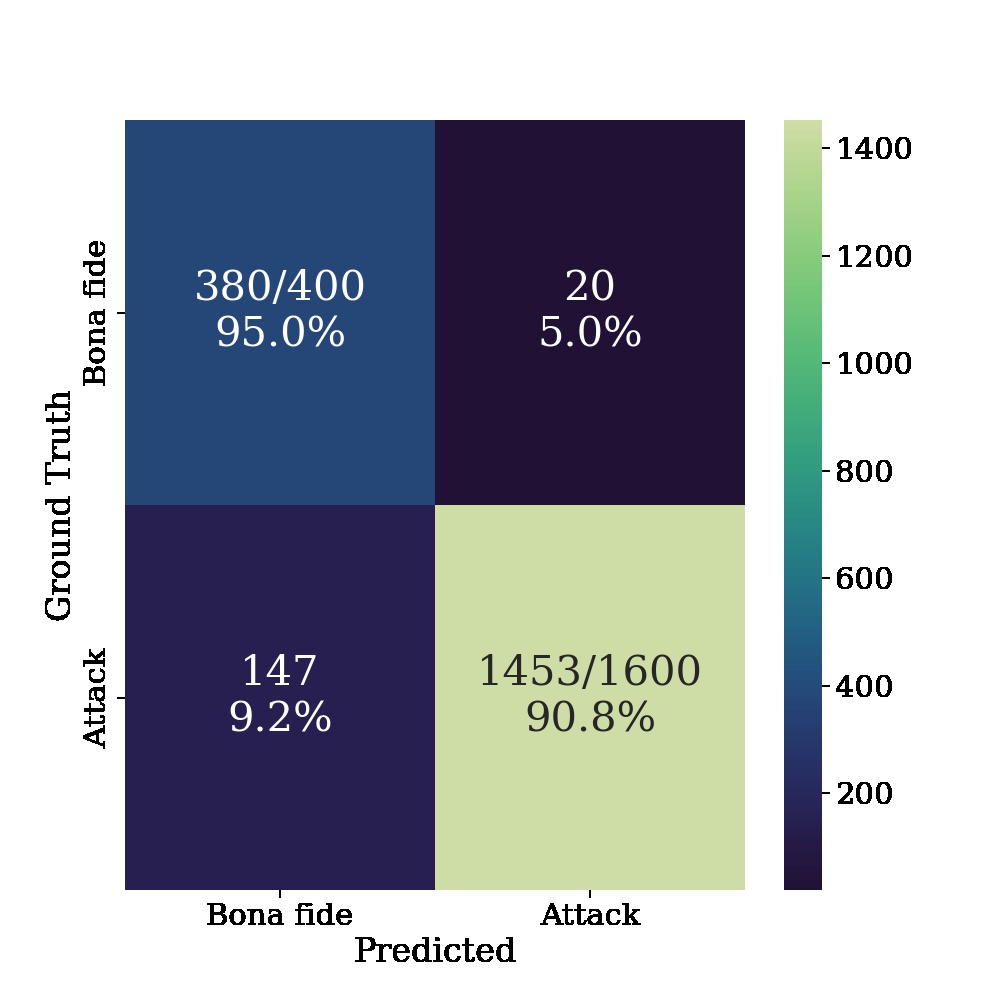}
\caption{DET curves for system trained with FRGCv2 and Tested with FERET for ResNet50+10 and MobileNetV2+15. Top: DET curves from ResNet50+10 and Confusion matrix. Bottom: DET curves from MobileNetV2+15 and Confusion matrix. The black dot line shows two operational point BPCER10 and BPCER20. In the parenthesis is show the D-EER in percentages for FaceFusion~(FF), FaceMorpher~(FM), Opencv-Morpher~(OCV) and UBO-Moprher~(UBO).}
\label{fig:det-exp3-res}
\end{figure*}

\subsection{Experiment 3}
This experiment was developed to explore the behaviour of the siamese network with FSL when adding a small number of examples from the target dataset, aiming to improve generalisation performance. For this purpose, 5, 10, 15 and 20 random samples per class, were removed from the test set (FERET) and included in the train set (FRGCv2), in order to study the effect in performance. Several optimisers were also explored in order to look for the best approach, obtaining the best results using Adam optimiser with a learning rate of $1e-5$. The best results of this Experiment are shown in Table \ref{tab:exp3}.

As shown in the first row of Table~\ref{tab:exp3} the network trained without any example from the target set, reached a BPCER10 of 16.60\%. Afterwards, several boosting shoot samples were included in the train set in order to guide the process of learning. This technique shows impressive improvements. Values of 5, 10, 15 and 20 random images were explored. 

For MobileNetV2, when the training set is complemented with only 5 images per class, the BPCER10 is reduced to 9.1\%. The best results were reached with only 15 images per class with a BPCER10 of 5.50\%. 

Regarding to ResNet50, as is shown in the sixth row of Table~\ref{tab:exp3} the network trained without any additional samples reached a BPCER10 of 21.10\%. We performed the same steps as in the case of MobileNetV2, adding 5, 10, 15 and 20 examples. For ResNet50, when the training set is complemented with only 5 images per class, the BPCER10 is reduced to 7.25\%. The best results were reached with only 10 images per class with a BPCER10 of 4.91\%. 
A summary of the results is shown in Table \ref{tab:exp3}.

\begin{table}[H]
\scriptsize
\centering
\caption{Results of the experiment 3. SHL represents, Semi Hard Loss. "+ N", represents the number of boosting examples added for FSL from FERET database.}
\label{tab:exp3}
\begin{tabular}{|c|cccc|}
\hline
\multirow{2}{*}{Model} & \multicolumn{4}{c|}{\begin{tabular}[c]{@{}c@{}}Train:FRGCv2 / \\ Test: FERET\end{tabular}}                                                                                                                                                   \\ \cline{2-5} 
                       & \multicolumn{1}{c|}{Loss} & \multicolumn{1}{c|}{\begin{tabular}[c]{@{}c@{}}D-EER\\ (\%)\end{tabular}} & \multicolumn{1}{c|}{\begin{tabular}[c]{@{}c@{}}BPCER10\\ (\%)\end{tabular}} & \begin{tabular}[c]{@{}c@{}}BPCER20\\ (\%)\end{tabular} \\ \hline
MobileNetV2            & \multicolumn{1}{c|}{SHL}  & \multicolumn{1}{c|}{18.69}& \multicolumn{1}{c|}{16.60}& 29.75 \\ \hline
MobileNetV2+5        & \multicolumn{1}{c|}{SHL}  & \multicolumn{1}{c|}{12.74} & \multicolumn{1}{c|}{10.00} & 21.99 \\ \hline
MobileNetV2+10       & \multicolumn{1}{c|}{SHL}  & \multicolumn{1}{c|}{9.75} & \multicolumn{1}{c|}{5.73}  & 15.50 \\ \hline
\textbf{MobileNetV2+15} & \multicolumn{1}{c|}{\textbf{SHL}}  & \multicolumn{1}{c|}{\textbf{6.75}}& \multicolumn{1}{c|}{\textbf{5.50}}                                                 & \textbf{9.25}                                                 \\ \hline
MobileNetV2+20       & \multicolumn{1}{c|}{SHL}  & \multicolumn{1}{c|}{9.93}                                               & \multicolumn{1}{c|}{9.58}                                                 & 16.25                                                 \\ \hline
ResNet50               & \multicolumn{1}{c|}{SHL}  & \multicolumn{1}{c|}{10.99}& \multicolumn{1}{c|}{21.10}& 43.00 \\ \hline
ResNet50+5           & \multicolumn{1}{c|}{SHL}  & \multicolumn{1}{c|}{8.00} & \multicolumn{1}{c|}{7.25} & 19.50 \\ \hline
\textbf{ResNet50+10} & \multicolumn{1}{c|}{\textbf{SHL}}  & \multicolumn{1}{c|}{\textbf{7.87}}& \multicolumn{1}{c|}{\textbf{4.91}}                                                 & \textbf{12.75}                                                 \\ \hline
ResNet50+15          & \multicolumn{1}{c|}{SHL}  & \multicolumn{1}{c|}{9.75}                                               & \multicolumn{1}{c|}{8.25}                                                 & 18.25                                                 \\ \hline
ResNet50+20          & \multicolumn{1}{c|}{SHL}  & \multicolumn{1}{c|}{9.85}                                               & \multicolumn{1}{c|}{8.10}                                                  & 15.25                                                 \\ \hline
\end{tabular}
\end{table}

Figure~\ref{fig:det-exp3-res} on top shows the DET curve and the confusion matrix for ResNet+10 (Table \ref{tab:exp3}). On bottom Figure \ref{fig:det-exp3-res} shows the DET curve and the confusion matrix for MobileNetV2+15 (Table \ref{tab:exp3}). In the parenthesis is show the D-EER in percentages for FaceFusion~(FF), FaceMorpher~(FM), Opencv-Morpher~(OCV) and UBO-Moprher~(UBO). The Confusion matrixes are described as bona fide and attacks~(FF, FM, OCV, and UBO together).

\section{Visualisation}
A t-SNE map projection was used to visualise the projection of the data to a 2D plot. This method shows non-linear connections in the data. The t-SNE algorithm calculates a similarity measure between pairs of instances in the high dimensional space and in the low dimensional space. It then tries to optimise these two similarity measures using a cost function.

\begin{figure}[]
\centering
	\includegraphics[scale=0.36]{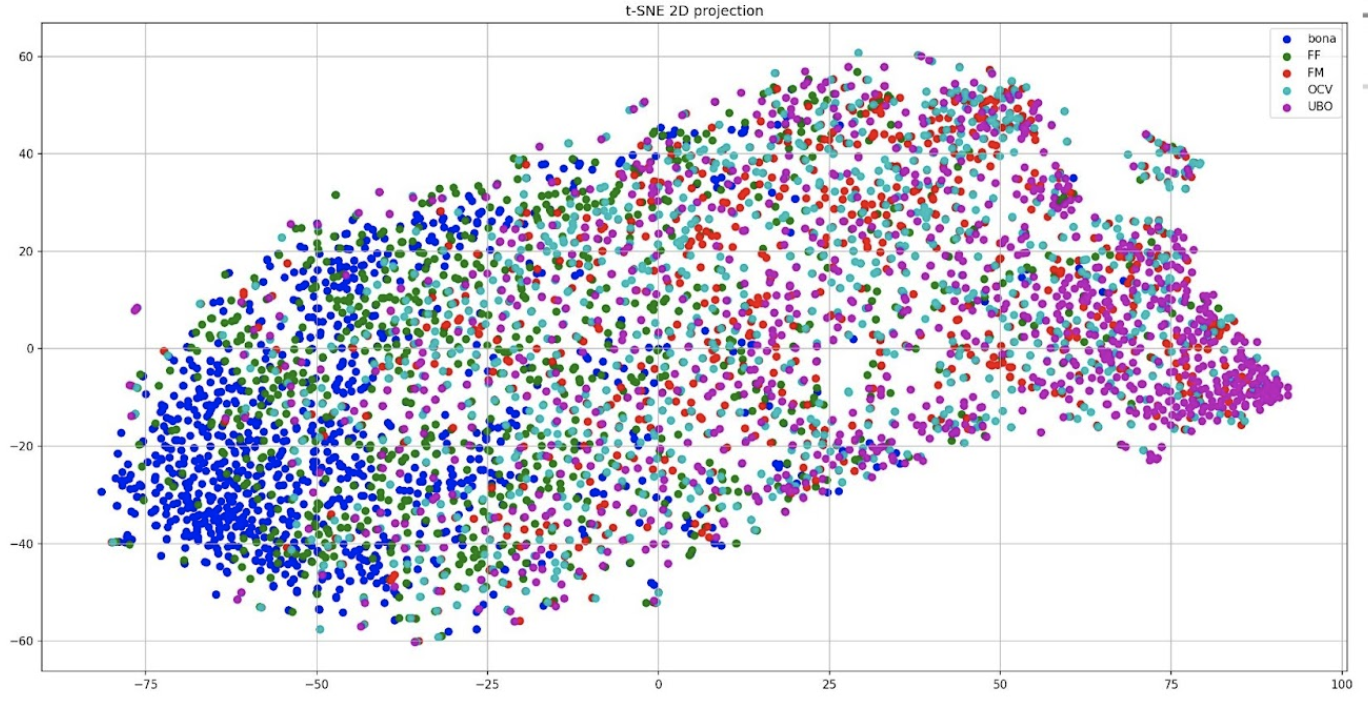}
\caption{t-SNE 2D projections. Blue color represents bona fide, Green is FaceFusion, Red is FaceMorpher, Lightblue is Opencv-Morpher, Purple represents UBO-Morpher. Top: Train FRGCvs and Tested with FERET. }
\label{fig:tsne-feret}
\end{figure}

Figure~\ref{fig:tsne-feret} shows the projection of FERET feature vectors, when the ResNet50 model with SHL was trained only with FRGCv2 samples Table~\ref{tab:exp1} columns 5 and 6. It can be observed that all classes overlaps, showing that the siamese network cannot separate bona fide from morphed examples. 

\begin{figure}[]
\centering
	\includegraphics[scale=0.35]{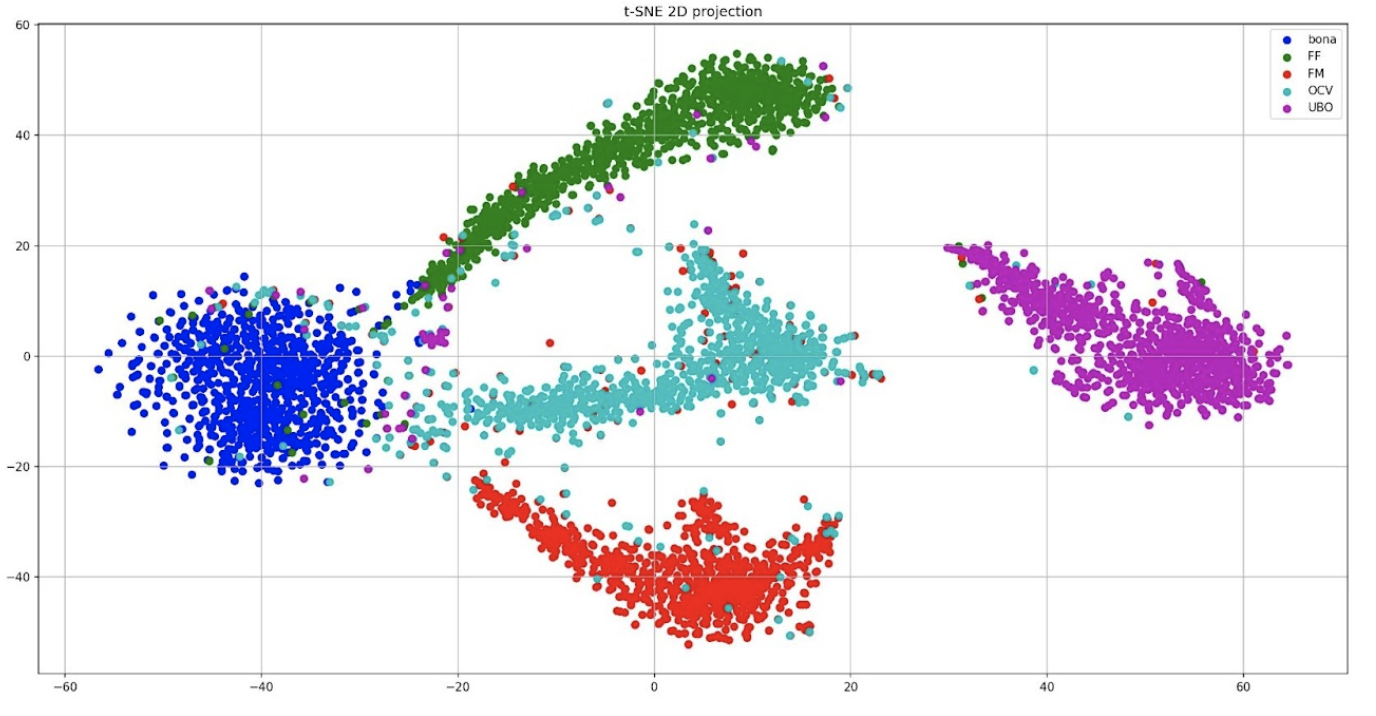}
\caption{t-SNE 2D projections. Blue color represents bona fide, Green is FaceFusion, Red is FaceMorpher, Lightblue is Opencv-Morpher, Purple represents UBO-Morpher. Train FRGCv2 and tested with MobileNetV2+15 few-shot per class. }
\label{fig:tsne-frgc}
\end{figure}

Figure~\ref{fig:tsne-frgc} shows the 2D projection when the FRGCv2 training set is complemented with only MobileNetV2 with 15 images per class from the FERET dataset (Table~\ref{tab:exp3}). It shows excellent separation and concentration of the all morphed classes and bona fide with MobileNetV2+15 (Table~\ref{tab:exp3}). Its important to mention that 15 images per class represents 0.5\% of FERET images.

\begin{figure}[]
\centering
	\includegraphics[scale=0.25]{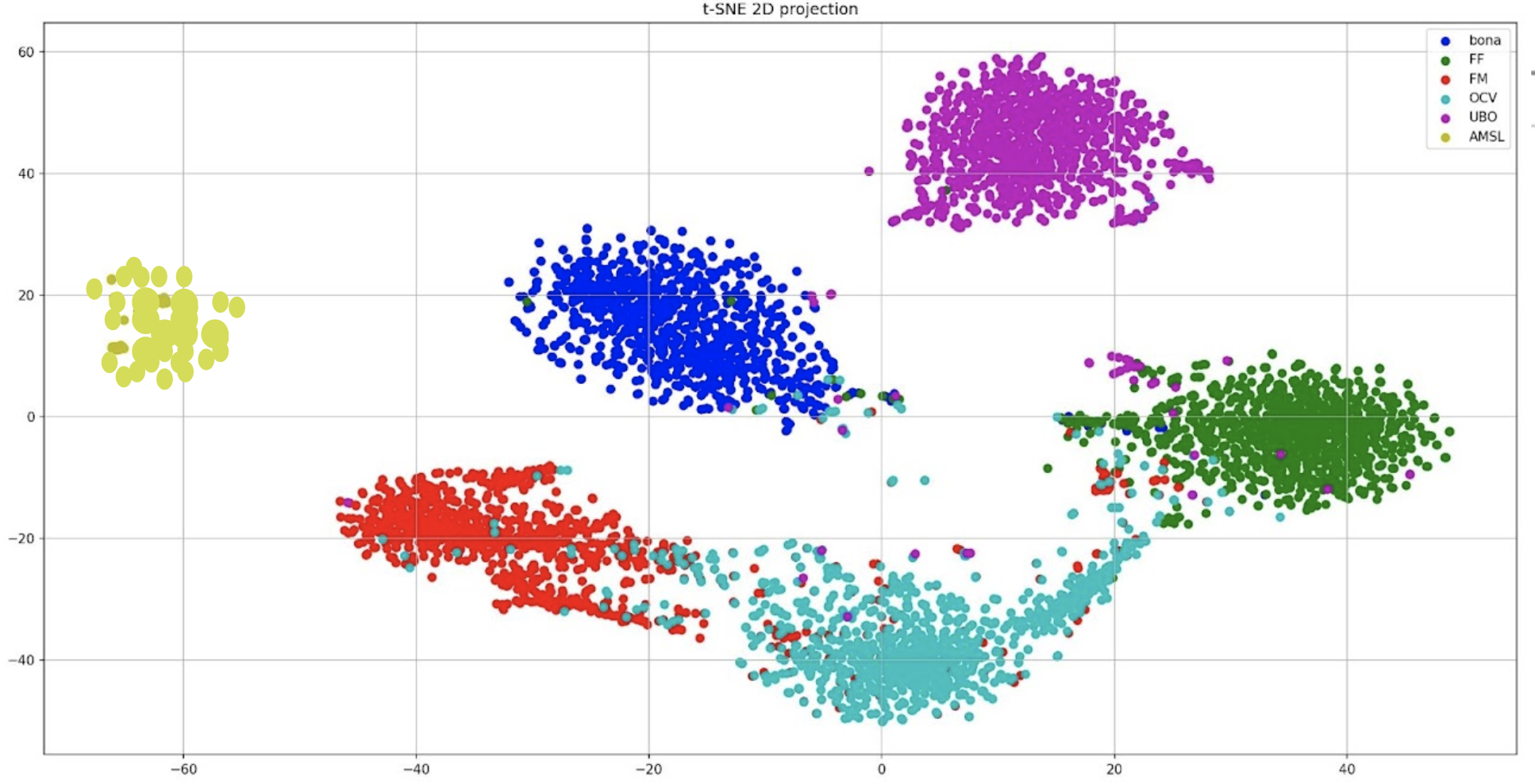}
\caption{t-SNE 2D projections. Blue color represents bona fide, Green is FaceFusion, Red is FaceMorpher, Lightblue is Opencv-Morpher, Purple represents UBO-Morpher. The new dataset AMSL was also included in yellow. }
\label{fig:tsne_amsl}
\end{figure}

Figure \ref{fig:tsne_amsl} shows the 2D projection when the FRGCv2 training set is complemented with only 15 images per class, applied to the morphed images from the AMSL open-access database. The yellow colour indicates the projection of AMSL morphed images into 2D space. It is essential to highlight that the new method allows us to identify new morphing tools. 

\section{Conclusion}
\label{conclusion}

The results reported in this paper show that including a few examples from the target domain, using the triplet loss semi-hard function, improve the results in a cross-database analysis. Results obtained adding only 15 samples from FERET to the FRGCv2 training dataset, gives a BPCER10 of 4.91\%.
A few examples from the new scenarios (morphing images) are enough to boost the learning process and get robust results. This approach is very plausible to be replicated when a new attack is detected in a simple operation. 
It is essential to highlight that we used ImageNet weights in ResNet50 and MobileNetV2. The ImageNet database has more than 1,000 classes of many objects, not only faces. Then, this proposed framework can be extended to other classification problems.

\section*{Acknowledgment}
This work is supported by the European Union’s Horizon 2020 research and innovation program under grant agreement No 883356 and the German Federal Ministry of Education and Research and the Hessen State Ministry for Higher Education, Research and the Arts within their joint support of the National Research Center for Applied Cybersecurity ATHENE and TOC Biometric R\&D SR-226 cooperation.

\section*{Disclaimer}
This text reflects only the author’s views, and the Commission is not liable for any use that may be made of the information contained therein.

\bibliographystyle{IEEEtran}
\bibliography{samples.bib}

\begin{IEEEbiography}[{\includegraphics[width=1in,height=1.25in,clip,keepaspectratio]{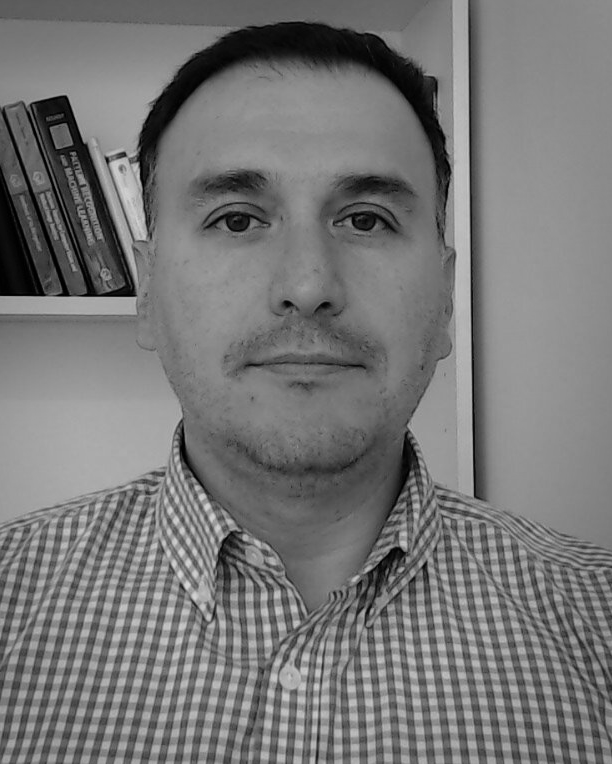}}]{Juan Tapia} received a P.E. degree in Electronics Engineering from Universidad Mayor in 2004, a M.S. in Electrical Engineering from Universidad de Chile in 2012, and a Ph.D. from the Department of Electrical Engineering, Universidad de Chile in 2016. In addition, he spent one year of internship at University of Notre Dame~(USA). In 2016, he received the award for best Ph.D. thesis. From 2016 to 2017, he was an Assistant Professor at Universidad Andres Bello. From 2018 to 2020, he was the R\&D Director for the area of Electricity and Electronics at Universidad Tecnologica de Chile - INACAP. He is currently a Senior Researcher at Hochschule Darmstadt~(HDA), and R\&D Director of TOC Biometrics. His main research interests include pattern recognition and deep learning applied to iris/face biometrics, vulnerability analysis, morphing, feature fusion, and feature selection.
\end{IEEEbiography}
\vspace{-1cm}

\begin{IEEEbiography}[{\includegraphics[width=1in,height=1.25in,clip,keepaspectratio]{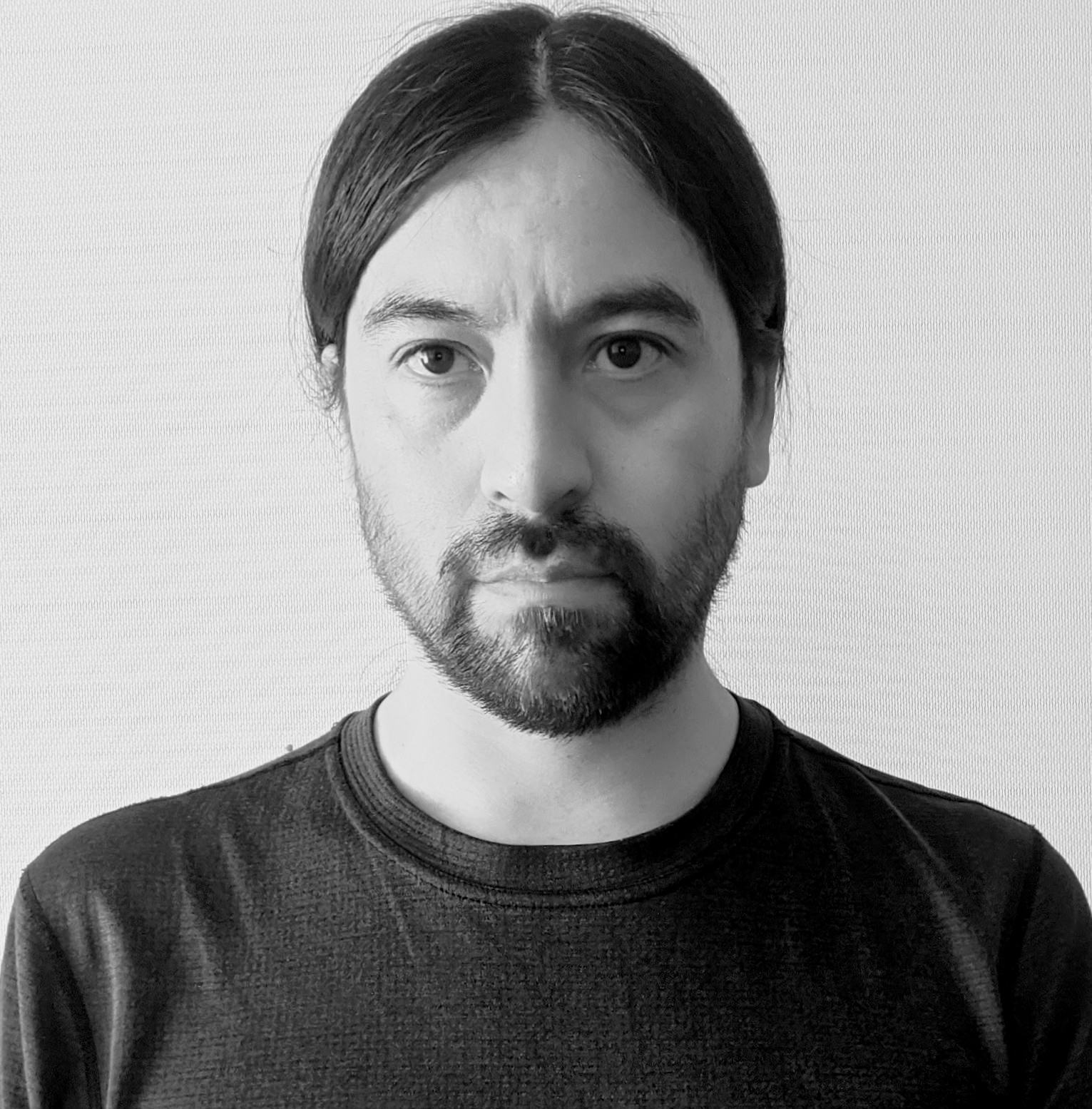}}]{Daniel Schulz} is a Ph.D. candidate from the Department of Electrical Engineering, Universidad de Chile, Santiago, Chile. He received the B.E. degree (Computer Science) from the Faculty of Engineering, Universidad Austral de Chile, in Valdivia, Chile, 2005. He is currently a Researcher at the R\&D center in TOC Biometrics. His main research interests are Biometrics, Computer Vision applied to Mining and Trademark Image Retrieval.
\end{IEEEbiography}

\begin{IEEEbiography}[{\includegraphics[width=1in,height=1.25in,clip,keepaspectratio]{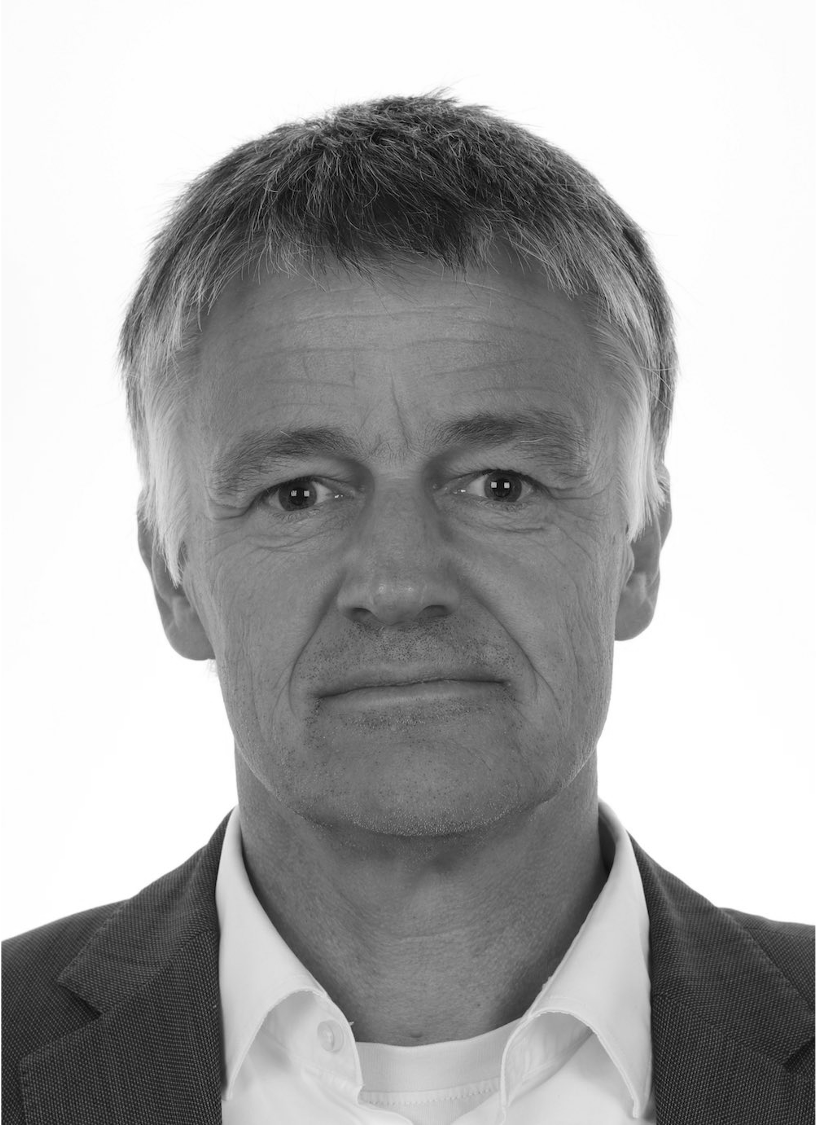}}]{Christoph Busch} is member of the Department of Information Security and Communication Technology (IIK) at the Norwegian University of Science and Technology (NTNU), Norway. He holds a joint appointment with the computer science faculty at Hochschule Darmstadt (HDA), Germany. Further he lectures the course Biometric Systems at Denmark’s DTU since 2007. On behalf of the German BSI he has been the coordinator for the project series BioIS, BioFace, BioFinger, BioKeyS Pilot-DB, KBEinweg and NFIQ2.0. In the European research program he was initiator of the Integrated Project 3D-Face, FIDELITY and iMARS. Further he was/is partner in the projects TURBINE, BEST Network, ORIGINS, INGRESS, PIDaaS, SOTAMD, RESPECT and TReSPAsS. He is also principal investigator in the German National Research Center for Applied Cybersecurity (ATHENE). Moreover Christoph Busch is co-founder and member of board of the European Association for Biometrics (www.eab.org) that was established in 2011 and assembles in the meantime more than 200 institutional members. Christoph co-authored more than 500 technical papers and has been a speaker at international conferences. He is member of the editorial board of the IET journal.
\end{IEEEbiography}

\end{document}